\documentclass[10pt,fleqn,twoside]{article}
\newcommand{\miniheader}{
  \documentclass[10pt,fleqn]{article}
  \usepackage{palatino}
  \usepackage{amsmath}
  \usepackage{amssymb}
  \usepackage{amsfonts}
  \usepackage{amsthm}
  \usepackage{eucal}
  \usepackage{graphicx}
  \usepackage{color}

  \usepackage{fancyhdr}
  \renewcommand{\headrulewidth}{.0pt}\renewcommand{\footrulewidth}{.0pt}\cfoot{}
  \fancyhead[OL,EC]{\it\theauthor---\today}
  \fancyhead[ER]{\leftmark}
  \fancyhead[OR,EL]{\thepage}
  \fancyfoot[EL,OR]{}

  \usepackage[round]{natbib}
  \bibliographystyle{abbrvnat}

  \graphicspath{{pics/}{figs/}{~/write/tex/pics/}{~/write/tex/figs/}}
  \usepackage{geometry}
  \geometry{a4paper,hdivide={35mm,*,35mm},vdivide={35mm,*,35mm}}
  \renewcommand{\baselinestretch}{1.1}

  \renewcommand{\a}{\alpha}
  \renewcommand{\b}{\beta}
  \renewcommand{\d}{\delta}
    \newcommand{\D}{\Delta}
    \newcommand{\e}{\epsilon}
    \newcommand{\g}{\gamma}
    \newcommand{\G}{\Gamma}
  \renewcommand{\l}{\lambda}
  \renewcommand{\L}{\Lambda}
    \newcommand{\m}{\mu}
    \newcommand{\n}{\nu}
    \newcommand{\N}{\nabla}
  \renewcommand{\k}{\kappa}
  \renewcommand{\o}{\omega}
  \renewcommand{\O}{\Omega}
    \newcommand{\p}{\varphi}
  \renewcommand{\P}{\Phi}
  \renewcommand{\r}{\varrho}
    \newcommand{\s}{\sigma}
  \renewcommand{\S}{\Sigma}
  \renewcommand{\t}{\theta}
    \newcommand{\T}{\Theta}
    \newcommand{\x}{\xi}
    \newcommand{\X}{\Xi}
    \newcommand{\Y}{\Upsilon}
    \newcommand{\z}{\zeta}

  \renewcommand{\AA}{{\cal A}}
    \newcommand{\BB}{{\cal B}}
    \newcommand{\CC}{{\cal C}}
    \newcommand{\cc}{{\cal c}}
    \newcommand{\DD}{{\cal D}}
    \newcommand{\EE}{{\cal E}}
    \newcommand{\FF}{{\cal F}}
    \newcommand{\GG}{{\cal G}}
    \newcommand{\HH}{{\cal H}}
    \newcommand{\II}{{\cal I}}
    \newcommand{\KK}{{\cal K}}
    \newcommand{\LL}{{\cal L}}
    \newcommand{\MM}{{\cal M}}
    \newcommand{\NN}{{\cal N}}
    \newcommand{\oNN}{\overline\NN}
    \newcommand{\OO}{{\cal O}}
    \newcommand{\PP}{{\cal P}}
    \newcommand{\QQ}{{\cal Q}}
    \newcommand{\RR}{{\cal R}}
  \renewcommand{\SS}{{\cal S}}
    \newcommand{\TT}{{\cal T}}
    \newcommand{\uu}{{\cal u}}
    \newcommand{\UU}{{\cal U}}
    \newcommand{\VV}{{\cal V}}
    \newcommand{\XX}{{\cal X}}
    \newcommand{\YY}{{\cal Y}}
    \newcommand{\SOSO}{{\cal SO}}
    \newcommand{\GLGL}{{\cal GL}}

    \newcommand{\Ee}{{\rm E}}

  \newcommand{\NNN}{{\mathbb{N}}}
  \newcommand{\III}{{\mathbb{I}}}
  \newcommand{\ZZZ}{{\mathbb{Z}}}
  \newcommand{\RRR}{{\mathbb{R}}}
  \newcommand{\SSS}{{\mathbb{S}}}
  \newcommand{\CCC}{{\mathbb{C}}}
  \newcommand{\DDD}{{\mathbb{D}}}
  \newcommand{\one}{{{\bf 1}}}
  \newcommand{\eee}{\text{e}}

  \newcommand{\NNNN}{{\overline{\cal N}}}

  \renewcommand{\[}{\Big[}
  \renewcommand{\]}{\Big]}
  \renewcommand{\(}{\Big(}
  \renewcommand{\)}{\Big)}
  \renewcommand{\|}{\,|\,}
  \renewcommand{\;}{\,;\,}
  \renewcommand{\=}{\!=\!}
    \newcommand{\<}{\left\langle}
  \renewcommand{\>}{\right\rangle}

  \newcommand{\na}{{\nabla\!}}
  \newcommand{\he}{{\nabla^2\!}}
  \newcommand{\Prob}{{\rm Prob}}
  \newcommand{\Dir}{{\rm Dir}}
  \newcommand{\Beta}{{\rm Beta}}
  \newcommand{\Var}{{\rm Var}}
  \newcommand{\Aut}{{\rm Aut}}
  \newcommand{\cor}{{\rm cor}}
  \newcommand{\corr}{{\rm corr}}
  \newcommand{\cov}{{\rm cov}}
  \newcommand{\sd}{{\rm sd}}
  \newcommand{\tr}{{\rm tr}}
  \newcommand{\Tr}{{\rm Tr}}
  \newcommand{\rank}{{\rm rank}}
  \newcommand{\diag}{{\rm diag}}
  \newcommand{\id}{{\rm id}}
  \newcommand{\Id}{{\rm\bf I}}
  \newcommand{\Gl}{{\rm Gl}}
  \renewcommand{\th}{\ensuremath{{}^\text{th}} }
  \newcommand{\lag}{\mathcal{L}}
  \newcommand{\inn}{\rfloor}
  \newcommand{\lie}{\pounds}
  \newcommand{\longto}{\longrightarrow}
  \newcommand{\speer}{\parbox{0.4ex}{\raisebox{0.8ex}{$\nearrow$}}}
  \renewcommand{\dag}{ {}^\dagger }
  \newcommand{\blbox}{\rule{1ex}{1ex}}
  \newcommand{\Ji}{J^\sharp}
  \newcommand{\h}{{}^\star}
  \newcommand{\w}{\wedge}
  \newcommand{\too}{\longrightarrow}
  \newcommand{\oot}{\longleftarrow}
  \newcommand{\To}{\Rightarrow}
  \newcommand{\oT}{\Leftarrow}
  \newcommand{\oTo}{\Leftrightarrow}
  \renewcommand{\iff}{~\Longleftrightarrow~}
  \newcommand{\Too}{\;\Longrightarrow\;}
  \newcommand{\oto}{\leftrightarrow}
  \newcommand{\ot}{\leftarrow}
  \newcommand{\ootoo}{\longleftrightarrow}
  \newcommand{\ow}{\stackrel{\circ}\wedge}
  \newcommand{\feed}{\nonumber \\}
  \newcommand{\comma}{~,\quad}
  \newcommand{\period}{~.\quad}
  \newcommand{\del}{\partial}
  \newcommand{\point}{$\bullet~~$}
  \newcommand{\doubletilde}{ ~ \raisebox{0.3ex}{$\widetilde {}$} \raisebox{0.6ex}{$\widetilde {}$} \!\! }
  \newcommand{\topcirc}{\parbox{0ex}{~\raisebox{2.5ex}{${}^\circ$}}}
  \newcommand{\topdot} {\parbox{0ex}{~\raisebox{2.5ex}{$\cdot$}}}
  \newcommand{\topddot} {\parbox{0ex}{~\raisebox{1.3ex}{$\ddot{~}$}}}
  \newcommand{\sym}{\topcirc}
  \newcommand{\tsum}{\textstyle\sum}
  \newcommand{\st}{\quad\text{s.t.}\quad}

  \newcommand{\half}{\ensuremath{\frac{1}{2}}}
  \newcommand{\third}{\ensuremath{\frac{1}{3}}}
  \newcommand{\fourth}{\ensuremath{\frac{1}{4}}}

  \newcommand{\ubar}{\underline}
  \renewcommand{\vec}{\boldsymbol}
  \renewcommand{\*}{\text{\footnotesize\raisebox{-.4ex}{*}{}}}

  \newcommand{\gto}{{\raisebox{.5ex}{${}_\rightarrow$}}}
  \newcommand{\gfrom}{{\raisebox{.5ex}{${}_\leftarrow$}}}
  \newcommand{\gnto}{{\raisebox{.5ex}{${}_\nrightarrow$}}}
  \newcommand{\gnfrom}{{\raisebox{.5ex}{${}_\nleftarrow$}}}

  \DeclareMathOperator*{\argmax}{argmax}
  \DeclareMathOperator*{\argmin}{argmin}
  \DeclareMathOperator{\sign}{sign}
  \DeclareMathOperator{\acos}{acos}
  \newcommand{\ee}[1]{\ensuremath{\cdot10^{#1}}}
  \newcommand{\sub}[1]{\ensuremath{_{\text{#1}}}}
  \newcommand{\up}[1]{\ensuremath{^{\text{#1}}}}
  \newcommand{\kld}[3][{}]{D_{#1}\big(#2\,\big|\!\big|\,#3\big)}
  \newcommand{\sprod}[2]{\big<#1\,,\,#2\big>}
  \newcommand{\End}{\text{End}}
  \newcommand{\txt}[1]{\quad\text{#1}\quad}
  \newcommand{\Over}[2]{\genfrac{}{}{0pt}{0}{#1}{#2}}
  \newcommand{\arr}[2]{\hspace*{-.5ex}\begin{array}{#1}#2\end{array}\hspace*{-.5ex}}
  \newcommand{\mat}[3][.9]{
    \renewcommand{\arraystretch}{#1}{\scriptscriptstyle{\left(
      \hspace*{-1ex}\begin{array}{#2}#3\end{array}\hspace*{-1ex}
    \right)}}\renewcommand{\arraystretch}{1.2}
  }
  \newcommand{\Mat}[3][.9]{
    \renewcommand{\arraystretch}{#1}{\scriptscriptstyle{\left[
      \hspace*{-1ex}\begin{array}{#2}#3\end{array}\hspace*{-1ex}
    \right]}}\renewcommand{\arraystretch}{1.2}
  }
  \newcommand{\case}[2][ll]{\left\{\arr{#1}{#2}\right.}
  \newcommand{\seq}[1]{\textsf{\<#1\>}}
  \newcommand{\seqq}[1]{\textsf{#1}}
  \newcommand{\floor}[1]{\lfloor#1\rfloor}
  \newcommand{\Exp}[2]{\text{E}_{#1}\{#2\}}
  \newcommand{\ex}{\setminus}

  \providecommand{\href}[2]{{\color{blue}USE PDFLATEX!}}
  \providecommand{\url}[2]{\href{#1}{{\color{blue}#2}}}
  \newcommand{\anchor}[2]{\begin{picture}(0,0)\put(#1){#2}\end{picture}}
  \newcommand{\pagebox}{\begin{picture}(0,0)\put(-3,-23){
    \textcolor[rgb]{.5,1,.5}{\framebox[\textwidth]{\rule[-\textheight]{0pt}{0pt}}}}
    \end{picture}}

  \newcommand{\hide}[1]{
    \begin{list}{}{\leftmargin0ex \rightmargin0ex \topsep0ex \parsep0ex}
       \helvetica{5}{1}{m}{n}
       \renewcommand{\section}{\par SECTION: }
       \renewcommand{\subsection}{\par SUBSECTION: }
       \item[$~~\blacktriangleright$]
       #1%$\blacktriangleleft~~$
       \message{^^JHIDE--Warning!^^J}
    \end{list}
  }
  \newcommand{\Hide}{\renewcommand{\hide}[1]{\message{^^JHIDE--Warning (hidden)!^^J}}}
  \newcommand{\HIDE}{\renewcommand{\hide}[1]{}}
  \newcommand{\fullhide}[1]{}
  \newcommand{\todo}[1]{{\tt[TODO: #1]}\message{^^JTODO--Warning: #1^^J}}
  \newcommand{\Todo}{\renewcommand{\todo}[1]{\message{^^JTODO--Warning (hidden)!^^J}}}
  \newcommand{\myauthor}[1]{\author{#1}\newcommand{\theauthor}{#1}}%\@author}
  \newcommand{\mytitle}[1]{\title{#1}\newcommand{\thetitle}{#1}}%\@title}
  \newcommand{\header}{\begin{document}\mytitle\cleardefs}
  \newcommand{\contents}{{\tableofcontents}\renewcommand{\contents}{}}
  \newcommand{\footer}{\small\bibliography{marc,bibs}\end{document}}
  \newcommand{\widepaper}{\usepackage{geometry}\geometry{a4paper,hdivide={25mm,*,25mm},vdivide={25mm,*,25mm}}}
  \newcommand{\moviex}[2]{\movie[externalviewer]{#1}{#2}} %\pdflatex\usepackage{multimedia}
  \newcommand{\rbox}[1]{\fboxrule2mm\fcolorbox[rgb]{1,.85,.85}{1,.85,.85}{#1}}
  \newcommand{\mpage}[2]{{\begin{minipage}{#1\columnwidth}#2\end{minipage}}}
  \newcommand{\redbox}[2]{\fboxrule1mm\fcolorbox[rgb]{1,.7,.7}{1,.7,.7}{\begin{minipage}{#1\columnwidth}\center#2\end{minipage}}}
  \newcommand{\onecol}[2]{
    \begin{minipage}[c]{#1\columnwidth}#2\end{minipage}}
  \newcommand{\twocol}[5][0]{
    \begin{minipage}[c]{#2\columnwidth}#4\end{minipage}\hspace*{#1\columnwidth}%
    \begin{minipage}[c]{#3\columnwidth}#5\end{minipage}}
  \newcommand{\threecol}[7][0]{%
    \begin{minipage}[c]{#2\columnwidth}#5\end{minipage}\hspace*{#1\columnwidth}%
    \begin{minipage}[c]{#3\columnwidth}#6\end{minipage}\hspace*{#1\columnwidth}%
    \begin{minipage}[c]{#4\columnwidth}#7\end{minipage}}
  \newcommand{\threecoltext}[7][c]{
    \begin{minipage}[#1]{#2\textwidth}#5\end{minipage}%
    \begin{minipage}[#1]{#3\textwidth}#6\end{minipage}%
    \begin{minipage}[#1]{#4\textwidth}#7\end{minipage}}
  \newcommand{\threecoltop}[7][0]{%
   \begin{minipage}[t]{#2\columnwidth}#5\end{minipage}\hspace*{#1\columnwidth}%
   \begin{minipage}[t]{#3\columnwidth}#6\end{minipage}\hspace*{#1\columnwidth}%
   \begin{minipage}[t]{#4\columnwidth}#7\end{minipage}}
  \newcommand{\fourcol}[9][0]{%
   \begin{minipage}[c]{#2\columnwidth}#6\end{minipage}\hspace*{#1\columnwidth}%
   \begin{minipage}[c]{#3\columnwidth}#7\end{minipage}\hspace*{#1\columnwidth}%
   \begin{minipage}[c]{#4\columnwidth}#8\end{minipage}\hspace*{#1\columnwidth}%
   \begin{minipage}[c]{#5\columnwidth}#9\end{minipage}}
  \newcommand{\helvetica}[4]{\setlength{\unitlength}{1pt}\fontsize{#1}{#1}\linespread{#2}\usefont{OT1}{phv}{#3}{#4}}
  \newcommand{\helve}[1]{\helvetica{#1}{1.5}{m}{n}}
  \newcommand{\german}{\usepackage[german]{babel}\usepackage[utf8]{inputenc}}

\newcommand{\norm}[1]{|\!|#1|\!|}
\newcommand{\expr}[1]{[\hspace{-.2ex}[#1]\hspace{-.2ex}]}

\newcommand{\Jwi}{J^\sharp_W}
\newcommand{\THi}{T^\sharp_H}
\newcommand{\Jci}{J^\natural_C}
\newcommand{\hJi}{{\bar J}^\sharp}
\renewcommand{\|}{\,|\,}
\renewcommand{\=}{\!=\!}
\newcommand{\myminus}{{\hspace*{-.0pt}\text{\rm -}\hspace*{-.5pt}}}
\newcommand{\myplus}{{\hspace*{-.0pt}\text{\rm +}\hspace*{-.5pt}}}
\newcommand{\1}{{\myminus1}}
\newcommand{\2}{{\myminus2}}
\newcommand{\3}{{\myminus3}}
\newcommand{\mT}{{\text{\rm -}\hspace*{-1pt}\top}}
\newcommand{\po}{{\myplus1}}
\newcommand{\pt}{{\myplus2}}
\renewcommand{\T}{{\!\top\!}}
\newcommand{\xT}{{\underline x}}
\newcommand{\uT}{{\underline u}}
\newcommand{\zT}{{\underline z}}
\newcommand{\Sum}{\textstyle\sum}
\newcommand{\Int}{\textstyle\int}
\newcommand{\Prod}{\textstyle\prod}

\newenvironment{centy}{
\vspace{15mm}
\large
\hspace*{5mm}
\begin{minipage}{8cm}\it\color{blue}
}{
\end{minipage}
}

\newcommand{\old}{{\text{old}}}
\newcommand{\new}{{\text{new}}}
\newcommand{\MAP}{{\text{MAP}}}
\newcommand{\ML}{{\text{ML}}}

\newcommand{\redArrow}{\quad\anchor{0,-1}{\includegraphics[scale=.5]{figs/redArrow}}}
\newcommand{\pub}[1]{{\color{green}\helvetica{8}{1.3}{m}{n}#1\\}}
\DeclareMathOperator{\opKL}{KL}
\newcommand{\KL}[2]{\opKL\big(#1\,\big|\!\big|\,#2\big)} %\left(#1 |\!| #2\right)}

\renewcommand{\show}[2][.8]{\centerline{\includegraphics[width=#1\columnwidth]{#2}}}
\newcommand{\showh}[2][.8]{\includegraphics[width=#1\columnwidth]{#2}}
\newcommand{\shows}[2][.8]{\centerline{\includegraphics[scale=#1]{#2}}}
\newcommand{\showhs}[2][.8]{\includegraphics[scale=#1]{#2}}
\newcommand{\mov}[2]{\movie[externalviewer]{{\color{blue}\small #1}}{movies/#2}}
\newcommand{\cen}[1]{\centerline{#1}}

\newcommand{\redoMacrosInProof}{
  \renewcommand{\d}{\delta}
  \renewcommand{\=}{\!=\!}
}

\makeatletter
\newenvironment{code}{\smallskip\newline%
  \begin{lrbox}{\@tempboxa}\begin{minipage}{.99\columnwidth}\helvetica{7}{1.1}{m}{n}
}{
  \end{minipage}\end{lrbox}%
  \colorbox[rgb]{.95,.95,.95}{\usebox{\@tempboxa}}\smallskip\newline
}\makeatother

\newcommand{\refeq}[1]{(\ref{#1})}

\usepackage{algorithm}
\usepackage{algpseudocode}
\algrenewcommand{\algorithmicrequire}{\textbf{Input:~~}}
\algrenewcommand{\algorithmicensure}{\textbf{Output:}}
\algrenewcommand{\algorithmiccomment}[1]{\qquad\hfill~\hspace*{-5ex}\textit{// #1}}
\algrenewcommand{\alglinenumber}[1]{\helvetica{6}{1.3}{m}{n}#1:}

\newenvironment{algo}[1][8]{
\quad\begin{minipage}{.8\columnwidth}\helvetica{#1}{1.3}{m}{n}
\medskip\hrule\medskip
\begin{algorithmic}[1]
}{
\end{algorithmic}
\medskip\hrule\medskip
\end{minipage}
}

}

\newcommand{\stdpackages}{
  \usepackage{amsmath}
  \usepackage{amssymb}
  \usepackage{amsfonts}
  \allowdisplaybreaks
  \usepackage{amsthm}
  \usepackage{eucal}
  \usepackage{graphicx}
  \usepackage{color}
  \usepackage{geometry}

  \usepackage{multicol} 
  \usepackage{fancyhdr}

  \newcommand{\draft}{\usepackage[light,first]{draftcopy}\draftcopyName{draft}{350}}
  \newcommand{\labels}{\usepackage{showlabels}}
  \newcommand{\maple}{\usepackage{maple2e}}
  \newcommand{\makeidx}{\usepackage{makeidx}\makeindex}
  \newcommand{\chicago}{\usepackage{chicago}\bibliographystyle{chicago}
    \renewcommand{\refname}{References\renewcommand{\refname}{}}}
  \newcommand{\natbib}{\usepackage[round]{natbib}\bibliographystyle{abbrvnat}}
  \newcommand{\showlines}{
    \usepackage[modulo]{lineno} %options: pagewise, modulo, mathlines
    \renewcommand{\BM}{\begin{linenomath}}
    \renewcommand{\EM}{\end{linenomath}}
    \linenumbers
    \modulolinenumbers[5]
  }\newcommand{\BM}{}\newcommand{\EM}{}
}

\newcommand{\pdflatex}{
  \definecolor{bluecol}{rgb}{0,0,.5}
  \definecolor{greencol}{rgb}{0,.4,0}
  \usepackage[
    colorlinks,
    urlcolor=bluecol,
    citecolor=black,
    linkcolor=bluecol,
    pdfborder={0 0 0},
    pdfpagemode=UseNone, %UseOutlines, %bookmarks are displayed by acrobat
    pdfauthor={Marc Toussaint}
  ]{hyperref}
  \DeclareGraphicsExtensions{.pdf,.png,.jpg,.eps}
  \renewcommand{\r}{\varrho}
  \renewcommand{\l}{\lambda}
  \renewcommand{\L}{\Lambda}
  \renewcommand{\s}{\sigma}
  \renewcommand{\b}{\beta}
  \renewcommand{\d}{\delta}
  \renewcommand{\k}{\kappa}
  \renewcommand{\t}{\tau}
  \renewcommand{\O}{\Omega}
  \renewcommand{\o}{\omega}
  \renewcommand{\SS}{{\cal S}}
  \renewcommand{\=}{\!=\!}
}
\newcommand{\stdtheorems}{
  \theoremstyle{plain}
  \newtheorem{theorem}{Theorem}
  \newtheorem{lemma}[theorem]{Lemma}
  \newtheorem{corollary}[theorem]{Corollary}
  \newtheorem{proposition}{Proposition}
  \newtheorem{conjecture}{Conjecture}
  \newtheorem{result}{Result}[section]
  \newtheorem{hypothesis}{Hypothesis}[section]
  \theoremstyle{definition}
  \newtheorem{definition}{Definition}
  \theoremstyle{remark}
  \newtheorem{remark}{Remark}[section]
  \newtheorem{example}{Example}[section]
  \newtheorem{algoTheo}{Algorithm}
  \newtheorem{testTheo}{Test}
}
\newcommand{\stdstyle}[1]{
  \stdpackages
  \stdtheorems
  \renewcommand{\labelenumi}{\textbf{(\roman{enumi})}}
  \renewcommand{\theenumi}{(\roman{enumi})} %for ref
  \newcommand{\itemdot}{\renewcommand{\labelitemi}{\bf $\cdot$}}
  \newcommand{\enumA}{\renewcommand{\labelenumi}{\textbf{\Alph{enumi}}}}
  \newcommand{\blockindent}{3ex}
  \renewcommand{\baselinestretch}{#1}
  \renewcommand{\arraystretch}{1.2}
  \renewcommand{\topfraction}{1}
  \renewcommand{\bottomfraction}{1}
  \renewcommand{\textfraction}{0}
  \columnsep 5ex
  \parindent 3ex
  \parskip 1ex

  \parindent 0pt
  \topsep 4pt plus 1pt minus 2pt
  \partopsep 1pt plus 0.5pt minus 0.5pt
  \itemsep 2pt plus 1pt minus 0.5pt
  \parsep 2pt plus 1pt minus 0.5pt
  \parskip .5pc %add _in_ {thebibliography} environment in *.bbl

  \setcounter{tocdepth}{3}
  \setcounter{secnumdepth}{3}

  \geometry{a4paper,hdivide={35mm,*,35mm},vdivide={35mm,*,35mm}}

  \renewcommand{\headrulewidth}{.0pt}\renewcommand{\footrulewidth}{.0pt}\cfoot{}
  \fancyhead[OL,EC]{\it\theauthor---\today}
  \fancyhead[ER]{\leftmark}
  \fancyhead[OR,EL]{\thepage}
  \fancyfoot[EL,OR]{}
  \setlength{\headsep}{10mm}
  \renewenvironment{abstract}
    {\vspace*{5ex}\begin{rblock}\hrule\vspace{1.5ex}{\bf Abstract.~}\small}
    {\vspace{2ex}\hrule\end{rblock}\vspace{5ex}}
  \newenvironment{keyword}
    {\par{\it Keywords:~}}
    {}
  \newcommand{\published}{}
  \def\makemytitle{%
    \thispagestyle{empty}
    \begin{list}{}{\leftmargin3ex \rightmargin3ex \topsep0ex \parsep0ex}\item[]
      \begin{center}
        {\fontsize{18}{25}\selectfont{\thetitle\\}}\vspace{5ex}

        {\fontsize{14}{16}\selectfont{\theauthor\\}}\vspace{1ex}

        {\footnotesize{\sl \addressFUB}\\ \emailBerlin}

        {\footnotesize \today}

        \vspace{1ex}
        {\small \published}
      \end{center}
    \end{list}
    \renewcommand{\maketitle}{\chapter{\thetitle}}
  }
}

\newcommand{\cleardefs}{
  \renewcommand{\article}[2]{}
  \renewcommand{\book}[2]{}
  \renewcommand{\draft}{}
  \renewcommand{\labels}{}
  \renewcommand{\maple}{}
  \renewcommand{\makeidx}{}
  \renewcommand{\chicago}{}
  \renewcommand{\pdflatex}{}
  \renewcommand{\header}{}
}

\newcommand{\article}[2]{
  \stdstyle{#2}
  \usepackage{palatino}
  
}

\newcommand{\lectureNote}{
  \documentclass[10pt,twocolumn,fleqn]{article}
  \usepackage{amsmath}
  \usepackage{amssymb}
  \usepackage{amsfonts}
  \usepackage{amsthm}
  \usepackage{eucal}
  \usepackage{graphicx}
  \usepackage{color}
  \usepackage{fancyhdr}
  \usepackage{geometry}
  \usepackage{palatino}

  \renewcommand{\baselinestretch}{1.1}
  \geometry{a4paper,headsep=7mm,hdivide={15mm,*,15mm},vdivide={20mm,*,15mm}}
    
  \allowdisplaybreaks

  \fancyhead[OL,ER]{\thetitle, \textit{Marc Toussaint}---\today}
  \fancyhead[C]{}
  \fancyhead[OR,EL]{\thepage}
  \fancyfoot{}
  \pagestyle{fancy}
  
}

\newcommand{\slideScript}{
}

\newcommand{\nips}{
  \documentclass{article}
  \usepackage{nips07submit_e,times}
  \stdpackages
  \pagestyle{plain}
}

\newcommand{\nipsben}{
  \documentclass{article}
  \usepackage{nips06}
  \stdpackages
  \pagestyle{plain}
}

\newcommand{\ijcnn}{
  \documentclass[10pt,twocolumn]{ijcnn}
  \stdpackages
  \bibliographystyle{abbrv} 
}

\newcommand{\springer}{
  \documentclass{springer_llncs}
  \renewcommand{\theenumi}{\alph{enumi}}
  \renewcommand{\labelenumi}{(\alph{enumi})}
  \renewcommand{\labelitemi}{$\bullet$}
  \bibliographystyle{abbrv}
  \stdpackages\stdtheorems
}

\newcommand{\elsevier}{
  \documentclass{elsart1p}
  \usepackage{natbib}
  \bibliographystyle{elsart-harv}
  \stdpackages
  \stdtheorems

}

\newcommand{\ieeejournal}{
  \documentclass[journal,twoside]{IEEEtran}
  \renewcommand{\theenumi}{\roman{enumi}}
  \renewcommand{\labelenumi}{(\roman{enumi})}
  \bibliographystyle{IEEEtran.bst}
  \usepackage{cite}
  \stdpackages
  \stdtheorems
  
}

\newcommand{\ieeeconf}{
  \documentclass[a4paper, 10pt, conference]{ieeeconf}
  \IEEEoverridecommandlockouts
  \overrideIEEEmargins
  \bibliographystyle{IEEEtran.bst}
  \stdpackages
  \stdtheorems
  \renewcommand{\theenumi}{\roman{enumi}}
  \renewcommand{\labelenumi}{(\roman{enumi})}
  
}

\newcommand{\foga}{
  \documentclass{article} 
  \stdpackages
  \usepackage{foga-02}
  \usepackage{chicago}
  \bibliographystyle{foga-chicago}
}

\newcommand{\book}[2]{
  \documentclass[#1pt,twoside,fleqn]{book}
  \newenvironment{abstract}{\begin{rblock}{\bf Abstract.~}\small}{\end{rblock}}
  \stdstyle{#2}

}

\newcommand{\letter}{

  \stdstyle{1.1}
  \usepackage{palatino}
  \renewcommand{\a}{\alpha}
  \renewcommand{\b}{\beta}
  \renewcommand{\d}{\delta}
    \newcommand{\D}{\Delta}
    \newcommand{\e}{\epsilon}
    \newcommand{\g}{\gamma}
    \newcommand{\G}{\Gamma}
  \renewcommand{\l}{\lambda}
  \renewcommand{\L}{\Lambda}
    \newcommand{\m}{\mu}
    \newcommand{\n}{\nu}
    \newcommand{\N}{\nabla}
  \renewcommand{\k}{\kappa}
  \renewcommand{\o}{\omega}
  \renewcommand{\O}{\Omega}
    \newcommand{\p}{\varphi}
  \renewcommand{\P}{\Phi}
  \renewcommand{\r}{\varrho}
    \newcommand{\s}{\sigma}
  \renewcommand{\S}{\Sigma}
  \renewcommand{\t}{\theta}
    \newcommand{\T}{\Theta}
    \newcommand{\x}{\xi}
    \newcommand{\X}{\Xi}
    \newcommand{\Y}{\Upsilon}
    \newcommand{\z}{\zeta}

  \renewcommand{\AA}{{\cal A}}
    \newcommand{\BB}{{\cal B}}
    \newcommand{\CC}{{\cal C}}
    \newcommand{\cc}{{\cal c}}
    \newcommand{\DD}{{\cal D}}
    \newcommand{\EE}{{\cal E}}
    \newcommand{\FF}{{\cal F}}
    \newcommand{\GG}{{\cal G}}
    \newcommand{\HH}{{\cal H}}
    \newcommand{\II}{{\cal I}}
    \newcommand{\KK}{{\cal K}}
    \newcommand{\LL}{{\cal L}}
    \newcommand{\MM}{{\cal M}}
    \newcommand{\NN}{{\cal N}}
    \newcommand{\oNN}{\overline\NN}
    \newcommand{\OO}{{\cal O}}
    \newcommand{\PP}{{\cal P}}
    \newcommand{\QQ}{{\cal Q}}
    \newcommand{\RR}{{\cal R}}
  \renewcommand{\SS}{{\cal S}}
    \newcommand{\TT}{{\cal T}}
    \newcommand{\uu}{{\cal u}}
    \newcommand{\UU}{{\cal U}}
    \newcommand{\VV}{{\cal V}}
    \newcommand{\XX}{{\cal X}}
    \newcommand{\YY}{{\cal Y}}
    \newcommand{\SOSO}{{\cal SO}}
    \newcommand{\GLGL}{{\cal GL}}

    \newcommand{\Ee}{{\rm E}}

  \newcommand{\NNN}{{\mathbb{N}}}
  \newcommand{\III}{{\mathbb{I}}}
  \newcommand{\ZZZ}{{\mathbb{Z}}}
  \newcommand{\RRR}{{\mathbb{R}}}
  \newcommand{\SSS}{{\mathbb{S}}}
  \newcommand{\CCC}{{\mathbb{C}}}
  \newcommand{\DDD}{{\mathbb{D}}}
  \newcommand{\one}{{{\bf 1}}}
  \newcommand{\eee}{\text{e}}

  \newcommand{\NNNN}{{\overline{\cal N}}}

  \renewcommand{\[}{\Big[}
  \renewcommand{\]}{\Big]}
  \renewcommand{\(}{\Big(}
  \renewcommand{\)}{\Big)}
  \renewcommand{\|}{\,|\,}
  \renewcommand{\;}{\,;\,}
  \renewcommand{\=}{\!=\!}
    \newcommand{\<}{\left\langle}
  \renewcommand{\>}{\right\rangle}

  \newcommand{\na}{{\nabla\!}}
  \newcommand{\he}{{\nabla^2\!}}
  \newcommand{\Prob}{{\rm Prob}}
  \newcommand{\Dir}{{\rm Dir}}
  \newcommand{\Beta}{{\rm Beta}}
  \newcommand{\Var}{{\rm Var}}
  \newcommand{\Aut}{{\rm Aut}}
  \newcommand{\cor}{{\rm cor}}
  \newcommand{\corr}{{\rm corr}}
  \newcommand{\cov}{{\rm cov}}
  \newcommand{\sd}{{\rm sd}}
  \newcommand{\tr}{{\rm tr}}
  \newcommand{\Tr}{{\rm Tr}}
  \newcommand{\rank}{{\rm rank}}
  \newcommand{\diag}{{\rm diag}}
  \newcommand{\id}{{\rm id}}
  \newcommand{\Id}{{\rm\bf I}}
  \newcommand{\Gl}{{\rm Gl}}
  \renewcommand{\th}{\ensuremath{{}^\text{th}} }
  \newcommand{\lag}{\mathcal{L}}
  \newcommand{\inn}{\rfloor}
  \newcommand{\lie}{\pounds}
  \newcommand{\longto}{\longrightarrow}
  \newcommand{\speer}{\parbox{0.4ex}{\raisebox{0.8ex}{$\nearrow$}}}
  \renewcommand{\dag}{ {}^\dagger }
  \newcommand{\blbox}{\rule{1ex}{1ex}}
  \newcommand{\Ji}{J^\sharp}
  \newcommand{\h}{{}^\star}
  \newcommand{\w}{\wedge}
  \newcommand{\too}{\longrightarrow}
  \newcommand{\oot}{\longleftarrow}
  \newcommand{\To}{\Rightarrow}
  \newcommand{\oT}{\Leftarrow}
  \newcommand{\oTo}{\Leftrightarrow}
  \renewcommand{\iff}{~\Longleftrightarrow~}
  \newcommand{\Too}{\;\Longrightarrow\;}
  \newcommand{\oto}{\leftrightarrow}
  \newcommand{\ot}{\leftarrow}
  \newcommand{\ootoo}{\longleftrightarrow}
  \newcommand{\ow}{\stackrel{\circ}\wedge}
  \newcommand{\feed}{\nonumber \\}
  \newcommand{\comma}{~,\quad}
  \newcommand{\period}{~.\quad}
  \newcommand{\del}{\partial}
  \newcommand{\point}{$\bullet~~$}
  \newcommand{\doubletilde}{ ~ \raisebox{0.3ex}{$\widetilde {}$} \raisebox{0.6ex}{$\widetilde {}$} \!\! }
  \newcommand{\topcirc}{\parbox{0ex}{~\raisebox{2.5ex}{${}^\circ$}}}
  \newcommand{\topdot} {\parbox{0ex}{~\raisebox{2.5ex}{$\cdot$}}}
  \newcommand{\topddot} {\parbox{0ex}{~\raisebox{1.3ex}{$\ddot{~}$}}}
  \newcommand{\sym}{\topcirc}
  \newcommand{\tsum}{\textstyle\sum}
  \newcommand{\st}{\quad\text{s.t.}\quad}

  \newcommand{\half}{\ensuremath{\frac{1}{2}}}
  \newcommand{\third}{\ensuremath{\frac{1}{3}}}
  \newcommand{\fourth}{\ensuremath{\frac{1}{4}}}

  \newcommand{\ubar}{\underline}
  \renewcommand{\vec}{\boldsymbol}
  \renewcommand{\*}{\text{\footnotesize\raisebox{-.4ex}{*}{}}}

  \newcommand{\gto}{{\raisebox{.5ex}{${}_\rightarrow$}}}
  \newcommand{\gfrom}{{\raisebox{.5ex}{${}_\leftarrow$}}}
  \newcommand{\gnto}{{\raisebox{.5ex}{${}_\nrightarrow$}}}
  \newcommand{\gnfrom}{{\raisebox{.5ex}{${}_\nleftarrow$}}}

  \DeclareMathOperator*{\argmax}{argmax}
  \DeclareMathOperator*{\argmin}{argmin}
  \DeclareMathOperator{\sign}{sign}
  \DeclareMathOperator{\acos}{acos}
  \newcommand{\ee}[1]{\ensuremath{\cdot10^{#1}}}
  \newcommand{\sub}[1]{\ensuremath{_{\text{#1}}}}
  \newcommand{\up}[1]{\ensuremath{^{\text{#1}}}}
  \newcommand{\kld}[3][{}]{D_{#1}\big(#2\,\big|\!\big|\,#3\big)}
  \newcommand{\sprod}[2]{\big<#1\,,\,#2\big>}
  \newcommand{\End}{\text{End}}
  \newcommand{\txt}[1]{\quad\text{#1}\quad}
  \newcommand{\Over}[2]{\genfrac{}{}{0pt}{0}{#1}{#2}}
  \newcommand{\arr}[2]{\hspace*{-.5ex}\begin{array}{#1}#2\end{array}\hspace*{-.5ex}}
  \newcommand{\mat}[3][.9]{
    \renewcommand{\arraystretch}{#1}{\scriptscriptstyle{\left(
      \hspace*{-1ex}\begin{array}{#2}#3\end{array}\hspace*{-1ex}
    \right)}}\renewcommand{\arraystretch}{1.2}
  }
  \newcommand{\Mat}[3][.9]{
    \renewcommand{\arraystretch}{#1}{\scriptscriptstyle{\left[
      \hspace*{-1ex}\begin{array}{#2}#3\end{array}\hspace*{-1ex}
    \right]}}\renewcommand{\arraystretch}{1.2}
  }
  \newcommand{\case}[2][ll]{\left\{\arr{#1}{#2}\right.}
  \newcommand{\seq}[1]{\textsf{\<#1\>}}
  \newcommand{\seqq}[1]{\textsf{#1}}
  \newcommand{\floor}[1]{\lfloor#1\rfloor}
  \newcommand{\Exp}[2]{\text{E}_{#1}\{#2\}}
  \newcommand{\ex}{\setminus}

  \providecommand{\href}[2]{{\color{blue}USE PDFLATEX!}}
  \providecommand{\url}[2]{\href{#1}{{\color{blue}#2}}}
  \newcommand{\anchor}[2]{\begin{picture}(0,0)\put(#1){#2}\end{picture}}
  \newcommand{\pagebox}{\begin{picture}(0,0)\put(-3,-23){
    \textcolor[rgb]{.5,1,.5}{\framebox[\textwidth]{\rule[-\textheight]{0pt}{0pt}}}}
    \end{picture}}

  \newcommand{\hide}[1]{
    \begin{list}{}{\leftmargin0ex \rightmargin0ex \topsep0ex \parsep0ex}
       \helvetica{5}{1}{m}{n}
       \renewcommand{\section}{\par SECTION: }
       \renewcommand{\subsection}{\par SUBSECTION: }
       \item[$~~\blacktriangleright$]
       #1%$\blacktriangleleft~~$
       \message{^^JHIDE--Warning!^^J}
    \end{list}
  }
  \newcommand{\Hide}{\renewcommand{\hide}[1]{\message{^^JHIDE--Warning (hidden)!^^J}}}
  \newcommand{\HIDE}{\renewcommand{\hide}[1]{}}
  \newcommand{\fullhide}[1]{}
  \newcommand{\todo}[1]{{\tt[TODO: #1]}\message{^^JTODO--Warning: #1^^J}}
  \newcommand{\Todo}{\renewcommand{\todo}[1]{\message{^^JTODO--Warning (hidden)!^^J}}}
  \newcommand{\myauthor}[1]{\author{#1}\newcommand{\theauthor}{#1}}%\@author}
  \newcommand{\mytitle}[1]{\title{#1}\newcommand{\thetitle}{#1}}%\@title}
  \newcommand{\header}{\begin{document}\mytitle\cleardefs}
  \newcommand{\contents}{{\tableofcontents}\renewcommand{\contents}{}}
  \newcommand{\footer}{\small\bibliography{marc,bibs}\end{document}}
  \newcommand{\widepaper}{\usepackage{geometry}\geometry{a4paper,hdivide={25mm,*,25mm},vdivide={25mm,*,25mm}}}
  \newcommand{\moviex}[2]{\movie[externalviewer]{#1}{#2}} %\pdflatex\usepackage{multimedia}
  \newcommand{\rbox}[1]{\fboxrule2mm\fcolorbox[rgb]{1,.85,.85}{1,.85,.85}{#1}}
  \newcommand{\mpage}[2]{{\begin{minipage}{#1\columnwidth}#2\end{minipage}}}
  \newcommand{\redbox}[2]{\fboxrule1mm\fcolorbox[rgb]{1,.7,.7}{1,.7,.7}{\begin{minipage}{#1\columnwidth}\center#2\end{minipage}}}
  \newcommand{\onecol}[2]{
    \begin{minipage}[c]{#1\columnwidth}#2\end{minipage}}
  \newcommand{\twocol}[5][0]{
    \begin{minipage}[c]{#2\columnwidth}#4\end{minipage}\hspace*{#1\columnwidth}%
    \begin{minipage}[c]{#3\columnwidth}#5\end{minipage}}
  \newcommand{\threecol}[7][0]{%
    \begin{minipage}[c]{#2\columnwidth}#5\end{minipage}\hspace*{#1\columnwidth}%
    \begin{minipage}[c]{#3\columnwidth}#6\end{minipage}\hspace*{#1\columnwidth}%
    \begin{minipage}[c]{#4\columnwidth}#7\end{minipage}}
  \newcommand{\threecoltext}[7][c]{
    \begin{minipage}[#1]{#2\textwidth}#5\end{minipage}%
    \begin{minipage}[#1]{#3\textwidth}#6\end{minipage}%
    \begin{minipage}[#1]{#4\textwidth}#7\end{minipage}}
  \newcommand{\threecoltop}[7][0]{%
   \begin{minipage}[t]{#2\columnwidth}#5\end{minipage}\hspace*{#1\columnwidth}%
   \begin{minipage}[t]{#3\columnwidth}#6\end{minipage}\hspace*{#1\columnwidth}%
   \begin{minipage}[t]{#4\columnwidth}#7\end{minipage}}
  \newcommand{\fourcol}[9][0]{%
   \begin{minipage}[c]{#2\columnwidth}#6\end{minipage}\hspace*{#1\columnwidth}%
   \begin{minipage}[c]{#3\columnwidth}#7\end{minipage}\hspace*{#1\columnwidth}%
   \begin{minipage}[c]{#4\columnwidth}#8\end{minipage}\hspace*{#1\columnwidth}%
   \begin{minipage}[c]{#5\columnwidth}#9\end{minipage}}
  \newcommand{\helvetica}[4]{\setlength{\unitlength}{1pt}\fontsize{#1}{#1}\linespread{#2}\usefont{OT1}{phv}{#3}{#4}}
  \newcommand{\helve}[1]{\helvetica{#1}{1.5}{m}{n}}
  \newcommand{\german}{\usepackage[german]{babel}\usepackage[utf8]{inputenc}}

\newcommand{\norm}[1]{|\!|#1|\!|}
\newcommand{\expr}[1]{[\hspace{-.2ex}[#1]\hspace{-.2ex}]}

\newcommand{\Jwi}{J^\sharp_W}
\newcommand{\THi}{T^\sharp_H}
\newcommand{\Jci}{J^\natural_C}
\newcommand{\hJi}{{\bar J}^\sharp}
\renewcommand{\|}{\,|\,}
\renewcommand{\=}{\!=\!}
\newcommand{\myminus}{{\hspace*{-.0pt}\text{\rm -}\hspace*{-.5pt}}}
\newcommand{\myplus}{{\hspace*{-.0pt}\text{\rm +}\hspace*{-.5pt}}}
\newcommand{\1}{{\myminus1}}
\newcommand{\2}{{\myminus2}}
\newcommand{\3}{{\myminus3}}
\newcommand{\mT}{{\text{\rm -}\hspace*{-1pt}\top}}
\newcommand{\po}{{\myplus1}}
\newcommand{\pt}{{\myplus2}}
\renewcommand{\T}{{\!\top\!}}
\newcommand{\xT}{{\underline x}}
\newcommand{\uT}{{\underline u}}
\newcommand{\zT}{{\underline z}}
\newcommand{\Sum}{\textstyle\sum}
\newcommand{\Int}{\textstyle\int}
\newcommand{\Prod}{\textstyle\prod}

\newenvironment{centy}{
\vspace{15mm}
\large
\hspace*{5mm}
\begin{minipage}{8cm}\it\color{blue}
}{
\end{minipage}
}

\newcommand{\old}{{\text{old}}}
\newcommand{\new}{{\text{new}}}
\newcommand{\MAP}{{\text{MAP}}}
\newcommand{\ML}{{\text{ML}}}

\newcommand{\redArrow}{\quad\anchor{0,-1}{\includegraphics[scale=.5]{figs/redArrow}}}
\newcommand{\pub}[1]{{\color{green}\helvetica{8}{1.3}{m}{n}#1\\}}
\DeclareMathOperator{\opKL}{KL}
\newcommand{\KL}[2]{\opKL\big(#1\,\big|\!\big|\,#2\big)} %\left(#1 |\!| #2\right)}

\renewcommand{\show}[2][.8]{\centerline{\includegraphics[width=#1\columnwidth]{#2}}}
\newcommand{\showh}[2][.8]{\includegraphics[width=#1\columnwidth]{#2}}
\newcommand{\shows}[2][.8]{\centerline{\includegraphics[scale=#1]{#2}}}
\newcommand{\showhs}[2][.8]{\includegraphics[scale=#1]{#2}}
\newcommand{\mov}[2]{\movie[externalviewer]{{\color{blue}\small #1}}{movies/#2}}
\newcommand{\cen}[1]{\centerline{#1}}

\newcommand{\redoMacrosInProof}{
  \renewcommand{\d}{\delta}
  \renewcommand{\=}{\!=\!}
}

\makeatletter
\newenvironment{code}{\smallskip\newline%
  \begin{lrbox}{\@tempboxa}\begin{minipage}{.99\columnwidth}\helvetica{7}{1.1}{m}{n}
}{
  \end{minipage}\end{lrbox}%
  \colorbox[rgb]{.95,.95,.95}{\usebox{\@tempboxa}}\smallskip\newline
}\makeatother

\newcommand{\refeq}[1]{(\ref{#1})}

\usepackage{algorithm}
\usepackage{algpseudocode}
\algrenewcommand{\algorithmicrequire}{\textbf{Input:~~}}
\algrenewcommand{\algorithmicensure}{\textbf{Output:}}
\algrenewcommand{\algorithmiccomment}[1]{\qquad\hfill~\hspace*{-5ex}\textit{// #1}}
\algrenewcommand{\alglinenumber}[1]{\helvetica{6}{1.3}{m}{n}#1:}

\newenvironment{algo}[1][8]{
\quad\begin{minipage}{.8\columnwidth}\helvetica{#1}{1.3}{m}{n}
\medskip\hrule\medskip
\begin{algorithmic}[1]
}{
\end{algorithmic}
\medskip\hrule\medskip
\end{minipage}
}

  \parskip2.5ex
  \pagestyle{plain}
  \renewcommand{\familydefault}{\sfdefault}
}

\newcommand{\letterhead}[3]{
  \thispagestyle{empty}
  \vspace*{10mm}
  \begin{minipage}[t]{8cm}
    #1
  \end{minipage}
  \hspace*{\fill}
  \begin{minipage}[t]{6cm}
    \mbox{}~\hfill Prof.\ Dr.\ Marc Toussaint\\
    \mbox{}~\hfill Freie Universit\"at Berlin\\
    \mbox{}~\hfill Arnimallee 7\\
    \mbox{}~\hfill 14195 Berlin, Germany\\
    \mbox{}~\hfill +49 30 838 52485\\
    \mbox{}~\hfill marc-toussaint@fu-berlin.de
  \end{minipage}

   \vspace*{5mm}\hfill #3, \today\\

   \vspace*{5mm}{\textbf{#2}}

   \vspace*{5mm}
}

\newcommand{\slides}{
  \newcommand{\thepage}{\arabic{mypage}}
  \documentclass[t,hyperref={bookmarks=true}]{beamer}
  \usetheme{default}
  \usefonttheme[onlymath]{serif}
  \setbeamertemplate{navigation symbols}{}
  \setbeamersize{text margin left=5mm}
  \setbeamersize{text margin right=5mm}
  \setbeamertemplate{itemize items}{{\color{black}$\bullet$}}

  \stdpackages
  \usepackage{multimedia}

  \definecolor{bluecol}{rgb}{0,0,.5}
  \definecolor{greencol}{rgb}{0,.6,0}
  \renewcommand{\arraystretch}{1.2}
  \columnsep 0mm

  \columnseprule 0pt
  \parindent 0ex
  \parskip 0ex
  \newcommand{\headerfont}{\helvetica{14}{1.5}{b}{n}}
  \newcommand{\slidefont} {\helvetica{10}{1.4}{m}{n}}
  \renewcommand{\small} {\helvetica{9}{1.4}{m}{n}}
  \renewcommand{\tiny} {\helvetica{8}{1.3}{m}{n}}

  \newcounter{mypage}
  \newcommand{\incpage}{\addtocounter{mypage}{1}\setcounter{page}{\arabic{mypage}}}
  \setcounter{mypage}{0}
  \resetcounteronoverlays{page}

  \pagestyle{fancy}
  \renewcommand{\headrulewidth}{0pt} %1pt}
  \renewcommand{\footrulewidth}{0pt} %.5pt}
  \cfoot{}
  \rhead{}
  \lhead{}
  \rfoot{~\anchor{-10,12}{\tiny\textsf{\arabic{mypage}/\pageref{lastpage}}}}
  \definecolor{grey}{rgb}{.8,.8,.8}
  \definecolor{head}{rgb}{.85,.9,.9}
  \definecolor{blue}{rgb}{.0,.0,.5}
  \definecolor{green}{rgb}{.0,.5,.0}
  \definecolor{red}{rgb}{.8,.0,.0}
  \newcommand{\inverted}{
    \definecolor{main}{rgb}{1,1,1}
    \color{main}
    \pagecolor[rgb]{.3,.3,.3}
  }
  
}

\newcommand{\titleslide}[4][Marc Toussaint]{
  \newcommand{\slideauthor}{#1}
  \newcommand{\slidevenue}{#3}
  \slidefont
  \incpage
  \begin{frame}
  \begin{center}
    \vspace*{15mm}

    {\headerfont #2\\}
        
    \vspace*{7mm}

    #1 \\

    \vspace*{5mm}

    {\small 
      Machine Learning \& Robotics Lab -- University of Stuttgart\\
      marc.toussaint@informatik.uni-stuttgart.de

      \vspace*{3mm}

      \emph{#3}
    }

    \vspace*{0mm}

  \end{center}
  \begin{itemize}\item[]~\\
    #4
  \end{itemize}
  \end{frame}
}

\newcommand{\titleslideempty}[3]{
  \slidefont
  \incpage
  \begin{frame}
  \begin{center}
    \vspace*{15mm}

    {\headerfont #1\\}
        
    \vspace*{5mm}

    {\small\emph{#2}} \\

  \end{center}
  \begin{itemize}\item[]~\\
    #3
  \end{itemize}
  \end{frame}
}

\newcommand{\oldslide}[2]{
  \slidefont
  \incpage\begin{frame}
  \footskip-5mm
  \setlength{\unitlength}{1mm}
  {\headerfont #1} \vspace*{-2ex}
  \begin{itemize}\item[]~\\
    #2
  \end{itemize}
  \end{frame}
}

\newcommand{\slide}[2]{
  \slidefont
  \incpage\begin{frame}
  \addcontentsline{toc}{section}{#1}
  \vfill
  {\headerfont #1} \vspace*{-2ex}
  \begin{itemize}\item[]~\\
    #2
  \end{itemize}
  \vfill
  \end{frame}
}

\newcommand{\slidetop}[2]{
  \slidefont
  \incpage\begin{frame}
  ~
  {\headerfont #1} \vspace*{-2ex}
  \begin{itemize}\item[]~\\
    #2
  \end{itemize}
  \end{frame}
}

\newcommand{\slideempty}[2]{
  \slidefont
  \incpage\begin{frame}
  \setlength{\unitlength}{1mm}
  \begin{picture}(0,0)(6,4)
  \put(0,0){{\color{head}\rule{130mm}{13mm}{}}}
  \end{picture}
  {\headerfont #1}\\[-2ex]
    #2
  \end{frame}
}

\newcommand{\sliden}[2]{
  \slidefont
  \incpage\begin{frame}
  \setlength{\unitlength}{1mm}
  \centerline{\headerfont #1}
  \vspace*{-2ex}
  \begin{enumerate}\item[]~\\
    #2
  \end{enumerate}
  \end{frame}
}

\newcommand{\slidetwo}[2]{
  \slidefont
  \incpage\begin{frame}
  \setlength{\unitlength}{1mm}
  \begin{picture}(0,0)(6,5)
  \put(0,0){{\color{head}\rule{130mm}{14mm}{}}}
  \end{picture}
  {\headerfont #1}\\[-2ex]
  \begin{multicols}{2}
  \begin{itemize}%\item[]~\\
    #2
  \end{itemize}
  \end{multicols}
  \end{frame}
}

\newcommand{\poster}{
  \documentclass[fleqn]{article}
  \stdpackages
  \renewcommand{\baselinestretch}{1}
  \renewcommand{\arraystretch}{1.8}

  \usepackage{geometry}
  \geometry{
    paperwidth=1189mm,
    paperheight=841mm, %841mm, %91.3cm, % 120cm
    headheight=0mm,
    headsep=0mm,
    footskip=1mm,
    hdivide={3cm,*,3cm},vdivide={1cm,*,1cm}}

  \setlength{\columnsep}{5cm}
  \columnseprule 3pt
  \renewcommand{\labelitemi}{\rule[.4ex]{.6ex}{.6ex}~}

  \pagestyle{empty}

  \definecolor{grey}{rgb}{.9,.9,.9}
  \newcommand{\inverted}{
    \definecolor{main}{rgb}{1,1,1}
    \color{main}
    \pagecolor[rgb]{.3,.3,.3}
  }

}

\author{Marc Toussaint}

\newcommand{\inilogo}[1][.25]{\includegraphics[scale=#1]{INI}}
\newcommand{\rublogo}[1][.25]{\includegraphics[scale=#1]{RUB}}
\newcommand{\edinlogo}[1][.25]{\includegraphics[scale=#1]{pics/eushield-fullcolour}}

\newcommand{\addressCologne}{
  Institute for Theoretical Physics\\
  University of Cologne\\
  50923 K\"oln, Germany\\
  {\tt mt@thp.uni-koeln.de}\\
  {\tt www.thp.uni-koeln.de/\~{}mt/}
}

\newcommand{\emailINI}{mt@neuroinformatik.ruhr-uni-bochum.de}
\newcommand{\phoneINI}{+49-234-32-27974}
\newcommand{\faxINI}{+49-234-32-14209}
\newcommand{\urlINI}{\texttt{www.neuroinformatik.rub.de/PEOPLE/mt/}}
\newcommand{\AddressINI}{
  Institut~f\"ur~Neuroinformatik,
  Ruhr-Universit\"at~Bochum, ND~04,
  44780~Bochum, Germany
}
\newcommand{\addressINI}{
  Institut~f\"ur~Neuroinformatik\\
  Ruhr-Universit\"at Bochum, ND~04\\
  44780~Bochum, Germany
}

\newcommand{\emailANC}{mtoussai@inf.ed.ac.uk}
\newcommand{\phoneANC}{+44 131 650 3089}
\newcommand{\faxANC}{+44 131 650 6899}
\newcommand{\urlANC}{homepages.inf.ed.ac.uk/mtoussai}
\newcommand{\AddressANC}{
  School of Informatics,\\
  Institute for Adaptive and Neural Computation,\\
  University of Edinburgh, 5 Forrest Hill,\\
  Edinburgh EH1 2QL, Scotland, UK
}
\newcommand{\addressANC}{
  School~of~Informatics,\\
  University~of~Edinburgh, 5~Forrest~Hill\\
  Edinburgh~EH1~2QL, Scotland,~UK
}

\newcommand{\AddressBerlin}{
  Machine Learning \& Robotics group\\
  TU Berlin\\
  Franklinstr. 28/29, FR 6-9\\
  10587 Berlin, Germany
}
\newcommand{\addressTUB}{
  Machine~Learning~\&~Robotics~group, TU~Berlin\\\small
  Franklinstr. 28/29,~FR~6-9, 10587~Berlin, Germany
}
\newcommand{\addressFUB}{
  Machine~Learning~\&~Robotics~lab, FU~Berlin\\\small
  Arnimallee 7, 14195~Berlin, Germany
}
\newcommand{\addressUSTT}{
  Machine~Learning~\&~Robotics~lab, U~Stuttgart\\\small
  Universit{\"a}tsstra{\ss}e 38, 70569~Stuttgart, Germany
}
\newcommand{\emailBerlin}{mtoussai@cs.tu-berlin.de}
\newcommand{\phoneBerlin}{+49 30 314 24470}

\newcommand{\phoneHonda}{+49-69-89011-717}
\newcommand{\AddressHonda}{
  Honda Research Institute Europe\\
  Carl-Legien-Strasse 30\\
  63073 Offenbach/Main
}
\newcommand{\addressHonda}{
  Honda~Research~Institute~Europe~GmbH,\\\small
  Carl-Legien-Strasse~30, 63073~Offenbach/Main
}

\newlength{\subsecwidth}

\newcommand{\subsec}[1]{
  \addtocontents{toc}{
    \protect\setlength{\subsecwidth}{\textwidth}\protect\addtolength{\subsecwidth}{-27ex}
      \protect\vspace*{-1.5ex}\protect\hspace*{20ex}
      \protect\begin{minipage}[t]{\subsecwidth}\protect\footnotesize\protect\textsf{#1}\protect\end{minipage}
      \protect\par
  }
  \begin{rblock}\it #1\end{rblock}\medskip\noindent
}
\newcommand{\tocsep}{
  \addtocontents{toc}{\protect\bigskip}
}
\newcommand{\Chapter}[1]{
\chapter*{#1}\thispagestyle{empty}
\addcontentsline{toc}{chapter}{\protect\numberline{}#1}
}
\newcommand{\Section}[1]{
  \section*{#1}
  \addcontentsline{toc}{section}{\protect\numberline{}#1}
}
\newcommand{\Subsection}[1]{
  \subsection*{#1}
  \addcontentsline{toc}{subsection}{\protect\numberline{}#1}
}
\newcommand{\content}[1]{
}
\newcommand{\sepline}[1][200]{
  \begin{center} \begin{picture}(#1,0)
    \line(1,0){#1}
  \end{picture}\end{center}
}
\newcommand{\sepstar}{
  \begin{center} {\vspace{0.5ex}\rule[1.2ex]{5ex}{.1pt}~*~\rule[1.2ex]{5ex}{.1pt}} \end{center}\vspace{-1.5ex}\noindent
}
\newcommand{\partsection}[1]{
  \vspace{5ex}
  \centerline{\sc\LARGE #1}
  \addtocontents{toc}{\contentsline{section}{{\sc #1}}{}}
}
\newcommand{\intro}[1]{\textbf{#1}\index{#1}}

\newcounter{parac}
\newcommand{\para}{\noindent\refstepcounter{parac}{\bf [{\roman{parac}}]}~~}
\newcommand{\Pref}[1]{[\emph{\ref{#1}}\,]}

\newenvironment{itemS}{
\par
\tiny
\begin{list}{--}{\leftmargin4ex \rightmargin0ex \labelsep1ex
  \labelwidth2ex \topsep0pt \parsep0ex \itemsep0pt}
}{
\end{list}
}

\newenvironment{items}{
\par
\small%fontsize{9}{9}\linespread{1.2}
\begin{list}{--}{\leftmargin4ex \rightmargin0ex \labelsep1ex \labelwidth2ex
\topsep0pt \parsep0ex \itemsep3pt}
}{
\end{list}
}

\newenvironment{block}[1][]{{\noindent\bf #1}
\begin{list}{}{\leftmargin\blockindent \topsep-\parskip}
\item[]
}{
\end{list}
}

\newenvironment{rblock}{
\begin{list}{}{\leftmargin\blockindent \rightmargin\blockindent \topsep-\parskip}\item[]}{\end{list}}

\newcounter{algoi}
\newenvironment{algoList}{
\begin{list}{{(\thealgoi)}}
{\usecounter{algoi} \leftmargin7ex \rightmargin3ex \labelsep1ex
  \labelwidth5ex \topsep-.5ex \parsep.5ex \itemsep0pt}
}{
\end{list}\vspace*{1ex}
}
\newenvironment{test}[1][]{
  \medskip\begin{testTheo}[#1]~\begin{algoList}
}{
  \end{algoList}\end{testTheo}
}

\newcounter{questi}
\newenvironment{question}{
\begin{list}{\textbf{\thealgoi.}}
{\usecounter{algoi} \leftmargin2ex \rightmargin0ex \labelsep1ex
  \labelwidth1ex \topsep0ex \parsep.5ex \itemsep0pt}
\addtocounter{questi}{1}
\item[\textsf{Q\thequesti:}]
}{
\end{list}
}

\newenvironment{colpage}{
\addtolength{\columnwidth}{-3ex}
\begin{minipage}{\columnwidth}
\vspace{.5ex}
}{
\vspace{.5ex}
\end{minipage}
}

\newenvironment{enum}{
\begin{list}{}{\leftmargin3ex \topsep0ex \itemsep0ex}
\item[\labelenumi]
}{
\end{list}
}

\newenvironment{cramp}{
\begin{quote} \begin{picture}(0,0)
        \put(-5,0){\line(1,0){20}}
        \put(-5,0){\line(0,-1){20}}
\end{picture}
}{
\begin{picture}(0,0)
        \put(-5,5){\line(1,0){20}}
        \put(-5,5){\line(0,1){20}}
\end{picture} \end{quote}
}

\newcommand{\localcite}[1]{{
\begin{bibunit}[chicago]
\renewcommand{\refname}{\vspace{-\parskip}} \let\chapter\phantom \let\section\phantom
\nocite{#1}
\putbib[bibs]
\end{bibunit}%use \setcounter{enumiv}{xx} in thebibliography environment
}}

\newcommand{\boxpage}[2][\textwidth]{
  \setcounter{equation}{0}
  \renewcommand{\theequation}{A.\arabic{equation}}
  \fboxsep1ex
  \fbox{
  \begin{minipage}{#1}
    #2
  \end{minipage}
  }
}

\newcommand{\tightmath}{
  \setlength{\jot}{0pt}
  \setlength{\abovedisplayskip}{.5ex}
  \setlength{\belowdisplayskip}{.5ex}
}

\newenvironment{myproof}{
  \small
  \noindent \textit{Proof.~}
  \tightmath
}{
  \hfill
  \begin{picture}(0,0)(0,0)
  \put(-3.5,12){\rule{5pt}{5pt}}
  \end{picture}
}

\newcommand{\lst}{
  \usepackage{listings}
  \lstset{ %
    language=C,                % choose the language of the code
    basicstyle=\normalfont\small,       % the size of the fonts that are used for the code
    frame=none,                   % adds a frame around the code
    tabsize=4,                      % sets default tabsize to 2 spaces
    captionpos=b,                   % sets the caption-position to bottom
    texcl=true,
    mathescape=true,
    escapechar=\#,
    columns=flexible,
    xleftmargin=6ex,
    numbers=left, numberstyle=\footnotesize, stepnumber=1, numbersep=3ex
  }
}

  \stdstyle{1.1}
  \usepackage{palatino}

\pdflatex
\usepackage{fancyvrb}
\DefineShortVerb{\@}
\fvset{numbers=left,xleftmargin=5ex}

\title{{\Huge\textsf{KOMO}}\\
Newton methods for $k$-order Markov Constrained Motion Problems\\
\href{http://ipvs.informatik.uni-stuttgart.de/mlr/marc/source-code/14-KOMO.tgz}{\textbf{\textsc{\normalsize
Download Source Code Here}}}
}

\begin{document}
\maketitle

\begin{abstract}
This is a documentation of a framework for robot motion optimization
that aims to draw on classical constrained optimization methods. With
one exception the underlying algorithms are classical ones:
Gauss-Newton (with adaptive stepsize and damping), Augmented
Lagrangian, log-barrier, etc. The exception is a novel any-time
version of the Augmented Lagrangian. The contribution of this
framework is to frame motion optimization problems in a way that makes
the application of these methods efficient, especially by defining a
very general class of robot motion problems while at the same time
introducing abstractions that directly reflect the API of the source
code.
\end{abstract}

\section{Introduction}

Let $x_t \in\RRR^n$ be a joint configuration and 
$x_{0:T} = (x_0,\ldots,x_T)$ a
trajectory of length $T$. Note that troughout this framework
we \emph{do not} represent trajectories in the phase space, where the
state is $(x_t, \dot x_t)$---we represent trajectories directly in
configuration space. We consider optimization problems of a general
``$k$-order non-linear sum-of-squares constrained'' form
\begin{align} \label{eqKOMO}
\min_{x_{0:T}}\quad&
\sum_{t=0}^{T} f_t(x_{t-k:t})^\T f_t(x_{t-k:t})
~+~ \sum_{t,t'} k(t,t') x_t^\T x_{t'} \feed
\st&
 \forall_t:~ g_t(x_{t-k:t}) \le 0\comma h_t(x_{t-k:t}) = 0 ~.
\end{align}
where $x_{t-k:t} = (x_{t-k},..,x_{t-1},x_t)$ are $k+1$ tuples of
consecutive states. The functions $f_t(x_{t-k:t}) \in \RRR^{d_t}$,
$g_t(x_{t-k:t})\in\RRR^{m_t}$, and $h_t(x_{t-k:t})\in\RRR^{l_t}$ are
arbitrary first-order differentiable non-linear $k$-order
vector-valued functions. These define cost terms or
inequality/equality constraints for each $t$.  Note that the first
cost vector $f_0(x_{-k},..,x_0)$ depends on states $x_t$ with negative
$t$. We call these $(x_{-k},..,x_{-1})$ the \emph{prefix}. The prefix
defines the initial condition of the robot, which could for instance
be resting at some given $x_0$. (A postfix to constrain the
endcondition in configuration space is optional.)

The term $k(t,t')$ is an optional kernel measuring the (desired)
correlation between time steps $t$ and $t'$, which we explored but in
practice hardly used.

The $k$-order cost vectors $f_t(x_{t-k:t}) \in \RRR^{d_t}$ are very
flexible in including various elements that can represent both
transition and task-related costs. This is detailed below. To give
first examples, for transitional costs we can penalize square
velocities using $k=1$ (depending on two consecutive configurations)
$f_t(x_{t\1}, x_t) = (x_t - x_{t\1})$, and square accelerations using
$k=2$ (depending on three consecutive configurations) $f_t(x_{t\2},
x_{t\1}, x_t) = (x_{t} + x_{t\2} - 2 x_{t\1})$.  Likewise, for larger
values of $k$, we can penalize higher-order finite-differencing
approximations of trajectory derivatives (e.g., jerk).  Moreover, for
$k=2$, using the equations of motion $M\ddot{x}_t + F = \tau_t$ with
$\ddot{x}_{t} \approx x_{t+1} + x_{t-1} - 2 x_{t}$, we can explicitly
penalize square torques using $f_t = \sqrt{H} M (x_t -2
x_{t\1} + x_{t\2}) + F)$, where $\sqrt{H}$ is the Cholesky
decomposition of a torque cost metric $H$, implying costs $f_t^\T f_t
= u_t^\T H u_t$.

%% For task costs, using terms of the form $f_t \supset (y^*-\phi(x))/\s$
%% we can induce squared potentials in some task space $\phi$. The
%% somewhat awkward notation $f_t \supset v$ means that we constructed
%% the vector $f_t$ by appending $v$ to $f_t$.

The inequality and equality constraints $g_t$ and $h_t$ are equally
general: we can impose $k$-order constraints on joint configuration
transitions (velocities, accelerations, torques) or in task spaces.

The optimization problem \refeq{eqKOMO} can be rewritten as
\begin{align}\label{eqGN}
\min_{x_{0:T}} f(x_{0:T})^\T f(x_{0:T}) \st g(x_{0:T}) \le 0\comma h(x_{0:T})=0
\end{align}
where $f=(f_0;..;f_T)$ is the concatenation of all $f_t$ and
$g=(g_0;..;g_T)$, $h=(h_0;..;h_T)$. This defines a constrained
sum-of-squares problem which lends to Gauss-Newton methods. Let
$J=\na_{x_{0:T}} \Phi$ be the global Jacobian. It is essential to
realize that the pseudo-Hessian $J^\T J$ (as used by Gauss-Newton) is
a \emph{banded} symmetric matrix. The band-width is $(k+1)n$.

\subsection{The KOMO code}

The goal of the implementation is the separation between the code of
optimizers and code to specify motion problems. The problem
form \refeq{eqKOMO} provides the abstraction for that interface. The
optimization methods all assume the general form
\begin{align}\label{eqOpt}
\min_x f(x) \st g(x)\le 0 \comma h(x) = 0
\end{align}
of a non-linear constrained optimization problem, with the additional
assumption that the (approximate) Hessian $\he f(x)$ can be provided
and is semi-pos-def. Therefore, the KOMO code essentially does the
following
\begin{itemize}
\item Provide interfaces to define sets of $k$-order task spaces and
costs/constraints in these task spaces at various time slices; which
constitutes a MotionProblem. Such a MotionProblem definition is very
semantic, referring to the kinematics of the robot.
\item Abstracts and converts a MotionProblem definition into the general
form \refeq{eqKOMO} using a kinematics engine. The resulting
MotionProblemFunction is not semantic anymore and provides the
interface to the generic optimization code.
\item Converts the problem definition \refeq{eqKOMO} into the general
forms \refeq{eqGN} and \refeq{eqOpt} using appropriate matrix packings
to exploit the chain structure of the problem. This code does not
refer to any robotics or kinematics anymore.
\item Applies various optimizers. This is generic code.
\end{itemize}

The code introduces specialized matrix packings to exploit the
structure of $J$ and to efficiently compute the banded matrix $J^\T
J$. Note that the rows of $J$ have at most $(k+1)n$ non-zero elements
since a row refers to exactly one task and depends only on one
specific tuple $(x_{t-k},..,x_t)$. Therefore, although $J$ is
generally a $D\times (T+1)n$ matrix (with $D=\sum_t \dim(f_t)$),
each row can be packed to store only $(k+1)n$ non-zero elements. We
introduced a \emph{row-shifted} matrix packing representation for
this. Using specialized methods to compute $J^\T J$ and $J^\T x$ for
any vector $x$ for the row-shifted packing, we can efficiently compute
the banded Hessian and any other terms we need in Gauss-Newton
methods.

\section{Formal problem representation}

The following definitions also document the API of the code.
\begin{description}
\item[KinematicEngine] is a mapping $\G:~ x \mapsto \G(x)$
that maps a joint configuration to a data structure $\G(x)$ which
allows to efficiently evaluate task maps. Typically $\G(x)$ stores
the frames of all bodies/shapes/objects and collision
information. More abstractly, $\G(x)$ is any data structure that is
sufficient to define the task maps below.

Note: In the code there is yet no abstraction KinematicEngine. Only
one specific engine (KinematicWorld) is used. It would be
straight-forward to introduce an abstraction for kinematic engines
pin-pointing exactly their role for defining task maps.

\item[TaskMap] is a mapping $\phi:~ (\G_{-k},..,\G_0) \mapsto
(y,J)$ which gets $k+1$ kinematic data structures as input and returns
some vector $y\in\RRR^d$ and its Jacobian $J\in\RRR(d\times n)$.

\item[Task] is a tuple $c=(\phi, \r_{0:T},
 y^*_{0:T},\textsf{mode})$ where $\phi$ is a TaskMap and the
parameters $\r_{0:T},y^*_{0:T} \in\RRR^{d\times T\po}$ allow for an
additional linear transformation in each time slice. Here,
$d=\dim(\phi)$ is the dimensionality of the task map. This defines the
transformed task map
\begin{align}
\hat\phi_t(x_{t-k},..,x_t)
& = \diag(\r_t) (\phi(\G(x_{t-k}),..,\G(x_t)) - y^*_t) ~,
\end{align}
which depending on $\textsf{mode}\in\{\textsf{cost, constraint}\}$ is
interpreted as cost or constraint term. Note that, in the cost case,
$y^*_{0:T}$ has the semantics of a reference target for the task
variable, and $\r^*_{0:T}$ of a precision. In the code,
$\r_{0:T},y^*_{0:T}$ may optionally be given as $1\times 1$, $1\times
T\po$, $d\times 1$, or $d\times T\po$ matrices---and an interpreted
constant along the missing dimensions.

\item[MotionProblem] is a tuple $(T,\CC,x_{-k:-1})$ which gives
the number of time steps, a list $\CC=\{c_i\}$ of Tasks, and
a \emph{prefix} $x_{-k:-1} \in\RRR^{k\times n}$. The prefix allows to
evaluate tasks also for time $t=0$, where the prefix defines the
kinematic configurations $\G(x_{-k}),..,\G(x_0)$ at negative
times.\footnote{Optionally one can set a postfix $x_{T+1:T+k}$ which
fixes the final condition.} This defines the optimization problem
\begin{align}
f(x_{0:T})
&= \sum_{t=0}^T f_t(x_{t-k:t})^\T f_t(x_{t-k:t})
   \st \forall_{t=0,..,T}:~ g_t(x_{t-k:t}) \le 0
\end{align}
Here, $f_t$ is the concatenation of all $\hat\phi_t^c$ over tasks
$c\in\CC:c.\text{mode=cost} \wedge c.\r_t\not=0$; and $g_t$ is the
concatenation of all $\hat\phi_t^c$ over tasks
$c\in\CC:c.\text{mode=constraint} \wedge c.\r_t\not=0$.

\end{description}

\section{User Interfaces}

\subsection{Easy}

For convenience there is a single high-level method to call the
optimization, defined in @<Motion/komo.h>@
\begin{code}
\begin{verbatim}
/// Return a trajectory that moves the endeffector to a desired target position
arr moveTo(ors::KinematicWorld& world, //in initial state
           ors::Shape& endeff,         //endeffector to be moved
           ors::Shape& target,         //target shape
           byte whichAxesToAlign=0,    //bit coded options to align axes
           uint iterate=1);            //usually the optimization methods may be called just
                                       //once; multiple calls -> safety
\end{verbatim}
\end{code}
The method returns an optimized joint space trajectory so that the
endeff reaches the target. Optionally the optimizer additionaly
aligns some axes between the coordinate frames. This is just one
typical use case; others would include constraining vector-alignments
to zero (orthogonal) instead of +1 (parallel), or directly specifying
quaternions, or using many other existing task maps. See expert
interface.

This interface specifies the relevant coordinate frames by referring
to Shapes. Shapes (@ors::Shape@) are rigidly attached to bodies
(``links'') and usually represent a (convex) collision
mesh/primitive. However, a Shape can also just be a marker frame
(@ShapeType markerST=5@), in which case it is just a convenience to
define reference frames attached to bodies. So, the best way to
determine the geometric parameters of the endeffector and target
(offsets, relative orientations etc) is by transforming the respective
shape frames (@Shape::rel@).

The method uses implicit parameters (grabbed from cfg file or command line or default):
\begin{code}
\begin{verbatim}
  double posPrec = MT::getParameter<double>("KOMO/moveTo/precision", 1e3);
  double colPrec = MT::getParameter<double>("KOMO/moveTo/collisionPrecision", -1e0);
  double margin = MT::getParameter<double>("KOMO/moveTo/collisionMargin", .1);
  double zeroVelPrec = MT::getParameter<double>("KOMO/moveTo/finalVelocityZeroPrecision", 1e1);
  double alignPrec = MT::getParameter<double>("KOMO/moveTo/alignPrecision", 1e3);
\end{verbatim}
\end{code}

\subsection{Expert using the included kinematics engine}

See the implementation of @moveTo@! This really is the core guide to
build your own cost functions.

More generically, if the user would like to implement new TaskMaps or
use some of the existing ones:
\begin{itemize}
\item The user can define new $k$-order task maps by instantiating the
abstraction. There exist a number of predefined task maps. The
specification of a task map usually has only a few parameters like
``which endeffector shape(s) are you referring to''. Typically, a good
convention is to define task maps in a way such that \emph{zero} is a
desired state or the constraint boundary, such as relative
coordinates, alignments or orientation. (But that is not necessary,
see the linear transformation below.)

\item To define an optimization problem, the user creates a list of
tasks, where each task is defined by a task map and parameters that
define how the map is interpreted as a) a cost term or b) an inequality
constraint. This interpretation allows: a linear
transformation separately for each $t$ (=setting a reference/target
and precision); how maps imply a constraint. This interpretation has a
significant number of parameters: for each time slice different
targets/precisions could be defined.
\end{itemize}

\subsection{Expert with own kinematics engine}

The code needs a data structure $\G(q_t)$ to represent the
(kinematic) state $q_t$, where coordinate frames of all
bodies/shapes/objects have been precomputed so that evaluation of task
maps is fast. Currently this is @KinematicWorld@.

Users that prefer using the own kinematics engine can instantiate the
abstraction. Note that the engine needs to fulfill two roles: it must
have a @setJointState@ method that also precomputes all frames of all
bodies/shapes/objects. And it must be siffucient as argument of your
task map instantiations.

\subsection{Optimizers}

The user can also only use the optimizers, directly instantiating the
$k$-order Markov problem abstraction; or, yet a level below, directly
instantiating the @ConstrainedProblem@ abstraction. Examples are given
in @examples/Optim/kOrderMarkov@ and
@examples/Optim/constrained@. Have a look at the specific
implementations of the benchmark problems, esp.\ the
@ParticleAroundWalls@ problem.

\subsection{Parameters \& Reporting}

Every run of the code generates a MT.log file, which tells about every
parameter that was internally used. You can overwrite any of these
parameters on command line or in an MT.cfg file.

Inspecting the cost report after an optimization is
important. Currently, the code goes through the task list $\CC$ and
reports for each the costs associated to it. There are also methods to
display the cost arising in the different tasks over time.

%% \section{Special Cases}

%% \subsection{``True'' dynamics for fully articulated systems}

%% %% penalize $u = M\ddot q + C\dot q + G$

%% %% Our kinematics: no efficient implementation of C!! Approximate $C\dot q
%% %% + G = F$ indep of $\dot q$; and all terms indep of $q$!!

%% \subsection{Jerk optimization: 3-order}

%% \subsection{Gaussian Process priors: kernel regularization}\label{secKernel}

%% \subsection{Inverse Kinematics: 1-order 1-step}

%% \subsection{Operational Space Control: 2-order 1-step}

%% \subsection{Endpose Optimization: 2-order 1-step}

%% \subsection{Multipose Optimization}

%% \section{Time Optimization}

\section{Potential Improvements}

There is many places the code code be improved (beyond documenting it
better):
\begin{items}
\item Implementing equality constraints: For a lack of necessity the
code does not yet handle equality constraints. We typically handle
equality tasks (reach a point) using cost terms; while focussing on
inequality constraints for collisions and joint limits.

\item The KinematicEngine should be abstracted to allow for easier
plugin of alternative engines.

\item Our kinematics engine uses SWIFT++ for proximity and penetration
computation. The methods would profit enormously from better (faster,
more accurate) proximity engines (signed distance functions, sphere-swept
primitives).
\end{items}

\section{Disclaimer}

This document by no means aims to document all aspects of the code,
esp.\ those relating to the used kinematics engine etc. It only tries
to introduce to the concepts and design decisions behind the KOMO
code.

More documentation of optimization and kinematics concepts used in the
code can be drawn from my teaching lectures on Optimization and
Robotics.

%\appendix

%\input{old2}
%\input{old}

\small
\bibliography{marc,theses,bibs}
\end{document}